\documentclass{article}
\usepackage{arxiv}
\usepackage[utf8]{inputenc} 
\usepackage[T1]{fontenc}    
\usepackage{hyperref}       
\usepackage{url}            
\usepackage{booktabs}       
\usepackage{amsfonts}       
\usepackage{nicefrac}       
\usepackage{microtype}      
\usepackage{graphicx}
\usepackage{natbib}
\usepackage{doi}
\usepackage{amsthm}
\usepackage{amsmath}
\usepackage{natbib}
\let\oldcite\cite
\renewcommand{\cite}[1]{\mbox{\oldcite{#1}}}
\let\oldcitep\citep
\renewcommand{\citep}[1]{\mbox{\oldcitep{#1}}}

\title{Unsupervised semantic segmentation of urban high-density multispectral point clouds}
\author{ \hspace{1mm}Oona Oinonen\thanks{Department of Remote Sensing and Photogrammetry, Finnish Geospatial Research Institute FGI, the National Land Survey of Finland, Vuorimiehentie 5, Espoo, FI-02150, Finland}\\
oona.oinonen@nls.fi
	\And
	\hspace{1mm}Lassi Ruoppa\footnotemark[1]
    \And
	\hspace{1mm}Josef Taher\footnotemark[1]
    \And
	\hspace{1mm}Matti Lehtomäki\footnotemark[1]
    \And
	\hspace{1mm}Leena Matikainen\footnotemark[1]
    \And
	\hspace{1mm}Kirsi Karila\footnotemark[1]
    \And
	\hspace{1mm}Teemu Hakala\footnotemark[1]
 \And
	\hspace{1mm}Antero Kukko\footnotemark[1]\hspace{2mm}\thanks{Department of Built Environment, School of Engineering, Aalto University, Aalto, FI-00076, Finland}
 \And
	\hspace{1mm}Harri Kaartinen\footnotemark[1]
 \And
	\hspace{1mm}Juha Hyyppä\footnotemark[1]
}

\renewcommand{\shorttitle}{Unsupervised semantic segmentation of urban high-density multispectral point clouds}

\hypersetup{
pdftitle={Unsupervised semantic segmentation of urban high-density multispectral point clouds},
pdfsubject={cs.CV},
pdfauthor={Oona Oinonen, Lassi Ruoppa, Josef Taher, Matti Lehtomäki, Leena Matikainen, Kirsi Karila, Teemu Hakala, Antero Kukko, Harri Kaartinen, Juha Hyyppä},
pdfkeywords={Multispectral point cloud, Unsupervised deep learning, Semantic segmentation, LiDAR, Airborne laser scanning (ALS), Land cover classification},
}

\begin{document}
\maketitle

\begin{abstract}
The availability of highly accurate urban airborne laser scanning (ALS) data will increase rapidly in the future, especially as acquisition costs decrease, for example through the use of drones. Current challenges in data processing are related to the limited spectral information and low point density of most ALS datasets. Another challenge will be the growing need for annotated training data, frequently produced by labour-intensive manual processes, to enable semantic interpretation of point clouds. This study proposes to semantically segment new high-density (1200 points per square metre on average) multispectral ALS data with an unsupervised ground-aware deep clustering method GroupSP inspired by the unsupervised GrowSP algorithm. GroupSP divides the scene into superpoints as a preprocessing step. The neural network is trained iteratively by grouping the superpoints and using the grouping assignments as pseudo-labels. The predictions for the unseen data are given by over-segmenting the test set and mapping the predicted classes into ground truth classes manually or with automated majority voting. GroupSP obtained an overall accuracy (oAcc) of 97\% and a mean intersection over union (mIoU) of 80\%. When compared to other unsupervised semantic segmentation methods, GroupSP outperformed GrowSP and non-deep K-means. However, a supervised random forest classifier outperformed GroupSP. The labelling efforts in GroupSP can be minimal; it was shown, that the GroupSP can semantically segment seven urban classes, i.e.,  building, high vegetation, low vegetation, asphalt, rock, football field, and gravel, with oAcc of 95\% and mIoU of 75\% using only 0.004\% of the available annotated points in the mapping assignment. Finally, the importance of the multispectral information was examined; adding each new spectral channel improved the mIoU. Additionally, echo deviation was valuable, especially when distinguishing ground-level classes.

\end{abstract}

\keywords{Multispectral point cloud \and Unsupervised deep learning \and Semantic segmentation \and LiDAR \and Airborne laser scanning (ALS) \and Land cover classification}

\section{Introduction}
\label{sec:introduction}

Three-dimensional (3D) airborne laser scanning (ALS) has become an indispensable source of
geospatial data, and it is used operationally for various mapping tasks, including nationwide
elevation mapping, 3D city modelling and monitoring of forest resources. The data is well suited
for automated analyses and can be used in many scene-understanding applications such as
semantic segmentation or classification tasks. The interest in such applications is great as the
demand for up-to-date and detailed information on built-up and natural environments is high and
the acquisition frequency and point density of ALS datasets is constantly increasing. Laser scanner
systems typically operate in one wavelength. The advantages of multispectral point clouds have
been shown, for example, in different land cover classification studies (e.g., \cite{wichmann2015evaluating}, \cite{matikainen2017object}, and \cite{teo2017analysis}). However, multispectral laser scanners are not yet
commonly available, and it is still important to study how to take full advantage of the multispectral
information \citep{zhang2022introducing}.

Similarly to the trend of object-based image analysis in remote sensing \citep{blaschke2010object}, segmentation of 3D scenes is desirable for multiple reasons as recognized, for example, in \cite{Poux2022AutomaticClouds}. For instance, scene processing and feature extraction are simplified if the operations can be done on a region rather than at a unit level. Flexible access to segments makes understanding the relations between neighbours, graphs and topology easier. Additionally, the point cloud annotation becomes more straightforward when operating on objects rather than points. The success of deep learning in computer vision and natural language processing has motivated active research on point cloud deep learning introducing new applications such as 3D semantic segmentation. However, the progress in 3D deep learning still falls behind the advantages of 2D computer vision.

An unsupervised approach to 3D semantic segmentation is extremely appealing, as it does not require annotated data often produced by a challenging and time-consuming manual process. Instead, the unsupervised methods use a pseudo target, for example, augmented instances or clustering assignment, to train the neural network (NN).
Unsupervised methods have shown great potential on point clouds, yet most studies concentrate on object-level rather than scene-level data \citep{Xiao2023UnsupervisedSurvey}.

High-density ALS point clouds are gaining increasing attention in urban scene applications due to their consistency in complex surface features \citep{wang2023semantic}.
However, unsupervised semantic segmentation of these point clouds is a great challenge given many open questions and problems \citep{wang2023semantic}. The feature extraction is complicated by data sparseness, data irregularity, noise, possible unseen objects in the new scanned areas, and the severe class imbalance typical in urban scenes.
Urban point clouds are collected on several square kilometres producing massive amounts of points. It is challenging to efficiently preprocess and feed a huge point cloud into a deep learning model with limited memory and computing capacity. Multispectral point clouds can provide improvements compared to conventional ALS datasets that suffer from limited spectral information. Nevertheless, only a few studies have focused on learning high-level semantic features directly from multispectral point clouds \citep{jing2021multispectral}. 

This work applies an unsupervised semantic segmentation deep learning pipeline to a high-density ALS multispectral point cloud to find seven urban object / land cover classes. 
Our contributions are listed as follows:
\begin{itemize}
    \item A new high-density multispectral ALS point cloud scanned from an urban area is proposed for semantic segmentation. The point cloud has an extremely high density of 1200 points per square metre on average; for example, the national laser scanning data from Finland contains 5 points per square metre at minimum \citep{nationalLaserScanning}.
    \item A novel concept is proposed to apply an unsupervised deep learning approach to such high-density multispectral data to reduce the manual annotation required in the semantic segmentation process. To our knowledge, such ALS data has not yet been attempted to semantically segment with unsupervised deep learning methods. 
    \item 
    For the semantic segmentation task, a novel ground-aware unsupervised deep clustering method GroupSP that groups similar superpoints (SP) using multispectral information and learned deep neural features is proposed. GroupSP is inspired by the pioneering GrowSP \citep{zhang2023growsp} algorithm that exploited similar techniques to segment indoor and autonomous driving datasets semantically. 
    GroupSP is compared to two other unsupervised models, i.e., GrowSP and K-means \citep{lloyd1982least}. The unsupervised results are benchmarked with a supervised machine learning algorithm random forest (RF) \citep{breiman2001random}. The labelling efficiency of GroupSP is critically reviewed to conclude if the future manual annotations needed in the surge of new ALS datasets could be reduced in comparison to the currently well-performing supervised methods.
    \item An ablation study examines the importance of spectral information by incrementally introducing more spectral channels for the GroupSP model.
\end{itemize}

\section{Related work}
\label{sec:relatedwork}

\subsection{Multispectral laser scanning}
\label{sec:multispectral_laser_scannning}
The role of spectral information in remote sensing has been studied extensively, for example, in \cite{shaw2003spectral}, \cite{schott2007remote}, and \cite{khan2018modern}. The interest in multispectral laser scanning is increasing steadily in many applications \citep{kaasalainen2019multispectral, kaasalainen2021hyperspectral} due to several motivating factors, such as damaged building surface inspection \citep{wehr2006multi}, examining the structural and physiological characteristics of forests \citep{morsdorf2009assessing, chen2010two, hakala2012full}, and enhanced distinction between rock types \citep{malkamaki2019portable}. 
Lately, there has been a trend in publishing intellectual properties related to multispectral laser scanning in autonomous driving, robotics, and surveying (e.g., \cite{velodyne2014multiwavepatent}, \cite{microvision2020multiwavepatent}, and \cite{mit2020multiwavepatent}) indicating a growing interest in commercialization and increasing the data availability.

In the field of airborne laser scanning, \cite{wang2014airborne} used dual-wavelength ALS data from two separate systems for land cover classification and achieved a clearly higher accuracy than with single-wavelength data. Later, data from the first operational multispectral ALS system Teledyne Optech Titan \citep{van2015first} have been used in many studies, further highlighting the importance of the spectral information in addition to the geometry of the objects and demonstrating potential applications of multispectral ALS. The applications include, for example, land cover classification (e.g., \cite{wichmann2015evaluating}, \cite{matikainen2017object}, \cite{teo2017analysis}, \cite{ekhtari2018classification}, and \cite{li2022agfp}), road mapping \citep{karila2017feasibility}, change detection for map updating \citep{matikainen2017object, matikainen2019toward} and tree species or tree type classification (e.g., \cite{yu2017single}, \cite{lindberg2021classification}, and \cite{enayetullah2022identifying}). It has been shown that the use of intensity information from three channels instead of one improved the results of land cover classification, and the multispectral information was particularly useful for distinguishing ground-level classes \citep{matikainen2017object, teo2017analysis}. The study by \cite{matikainen2017object} suggested that multispectral ALS data has clear benefits compared to passive aerial images and has a high potential for further increasing the automation level in mapping. \cite{karila2017feasibility} investigated the topic further and reported promising results in recognising gravel roads and paved roads from multispectral ALS data. \cite{matikainen2019toward} tested change detection between two multispectral ALS datasets. Instead of Optech Titan data, \cite{hakula2023individual} used novel, high-density multispectral data from three separate scanners for tree species classification. The best results were obtained by using multispectral reflectance and geometric features.

Several multispectral land cover studies have used machine learning, and the number of deep learning approaches has increased in recent years. Some methods have preprocessed the point clouds into rasters or 2D images \citep{yu2019hybrid, pan2020land, yu2022capvit}. 
\cite{wang2021multi}, \cite{zhao2021airborne}, \cite{li2022agfp}, and \cite{yang2023multiscale} applied graph convolutional networks directly on multispectral point clouds.
Transformer networks have also been adapted for multispectral land cover classification \citep{zhang2022introducing}.
\cite{chen2024feature} applied a weakly supervised method for the classification of multispectral ALS point clouds by using  0.1\% of labelled points while obtaining mean intersection over union (mIoU) of 79\% with six classes.

\subsection{Deep learning for ALS data}

The architectures for point cloud deep learning have made remarkable progress during the last few years \citep{Xiao2023UnsupervisedSurvey}. PointNet \citep{qi2017pointnet} was the first deep architecture that works directly on points, and PointNet++ \citep{Qi2017PointNet++:Space} was shortly suggested to improve PointNet by learning features from the point's neighbourhood. 
An efficient method for large point clouds is RandLA-Net \citep{hu2020randla}, which is based on random sampling and applied for ALS classification, for example, in \cite{he2022offs}. Dynamic Graph Convolutional Neural Network (DGCNN) \citep{Wang2019DynamicClouds} performs convolutions on point cloud graphs and is suitable for ALS semantic segmentation \citep{wicaksono2019semantic}. Voxel-based networks divide the point cloud into voxels before applying 3D convolutions. Unnecessary computations on empty voxels can be avoided with sparse operations, for example, the sparse convolution library MinkowskiEngine \citep{Choy20194DNetworks}. Lately, transformer networks have been explored in point cloud processing after their success in natural language and 2D image processing \citep{Xiao2023UnsupervisedSurvey}.

Several studies have concentrated on proposing neural networks for ALS point clouds for applications such as semantic segmentation (e.g., \cite{huang2021granet}, \cite{lin2021local}, \cite{liu2022context}, \cite{guo2023mctnet}, \cite{zeng2023multi}, \cite{zeng2023recurrent}, and \cite{zhang2023pointboost}) and classification (e.g., \cite{wen2020directionally} and \cite{wen2021airborne}). 

\subsection{Reduced labelling efforts in ALS semantic segmentation}

Nonsupervised methods reduce the labelling efforts required in fully supervised approaches.
In \cite{hosseiny2023beyond}, the nonsupervised methods applied to remote sensing data are divided into the following categories. 

Transfer learning uses data from different domains to train and evaluate the model. \cite{zhao2019point} proposed transfer learning for classifying ALS point clouds, and \cite{huang2024simple} suggested few-shot learning for ALS semantic segmentation.

Weakly supervised methods seek to produce highly generalizable models by using inaccurate or inexact training data \citep{hosseiny2023beyond}. For ALS semantic segmentation, \cite{wang2022new} proposed training the network with only a fraction of labelled samples and \cite{lin2022weakly} exploited weak annotations on a group of points. 

Semi-supervised learning methods generate and select pseudo-labels for the test samples by some given criterion \citep{hosseiny2023beyond}. A subcategory, active learning samples the most informative points for a human annotator and was adapted, for example, in \cite{lin2020active} for ALS semantic segmentation.

Self-supervised learning pretrains the model in an unsupervised manner and the weights are fine-tuned for the main application. Self-supervision was applied for ALS semantic segmentation, for example, in \cite{zhang2024havana} and \cite{caros2023self}.

\subsection{Unsupervised deep learning for scene-level tasks}

In addition to the reduced labelling efforts, some studies propose fully unsupervised approaches for scene-level tasks exploiting deep learning.
A typical approach uses deep clustering techniques, i.e., neural features are clustered to produce pseudo-labels for training.

GrowSP \citep{zhang2023growsp} and U3DS3 \citep{liu2024u3ds3} are scene-level fully unsupervised semantic segmentation methods. The scene is oversegmented into spatially homogeneous superpoints. Superpoints are clustered using the learned deep features to produce pseudo-labels for the training data.
Similarly, PointDC \citep{chen2023pointdc} feeds multiview images of the point cloud to a pre-trained 2D network to calculate features that are projected back to 3D and clustered for pseudo-labels.  
In \cite{zhang2021unsupervised}, deep clustering is applied to predict labels for ALS scenes covering a small area and in \cite{de2023dc3dcd, de2023deep} for ALS change detection.
OGC \citep{song2022ogc} was proposed for unsupervised object segmentation for indoor and outdoor scenes.

\section{Materials and methods}
\label{sec:materialsandmethods}
\subsection{Dataset collection}
\label{sec:dataset_collection}
\begin{figure}
    \centering
    \includegraphics[width=14cm]{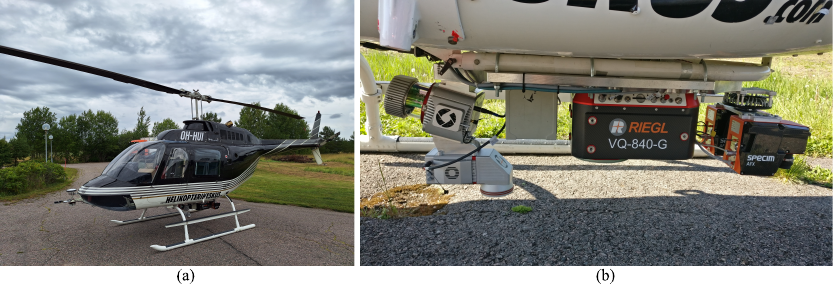}
    \caption{FGI's in-house developed HeliALS-TW (a). The high-density multispectral ALS data was collected with three LiDAR scanners mounted on a helicopter (b).}
    \label{fig:lidar_setup}
\end{figure}

The high-density multispectral point cloud data was collected from Espoonlahti (centre point approx. 60\textdegree 08'50.6"N 24\textdegree 39'17.7"E) in the south-western part of the capital region area of Helsinki, Finland in the summer of 2023. HeliALS-TW is a Finnish Geospatial Research Institute's (FGI) in-house developed multispectral laser scanning system mounted on a helicopter and used in \cite{hakula2023individual}. The system involves three Riegl scanners (Riegl GmbH, Austria): VUX-1HA (scanner 1), miniVUX-1DL (scanner 2), and VQ-840-G (scanner 3) with wavelengths of 1550 nm (infrared), 905 nm (near-infrared), and 532 nm (green), respectively. 
The positioning system comprised a NovAtel (LITEF) ISA-100C inertial measurement unit, a NovAtel PwrPak7 GNSS receiver, and a NovAtel (Vexxis) GNSS-850 antenna. The setup is shown in Figure~\ref{fig:lidar_setup}.

The data was collected by flying a helicopter over the test site from two perpendicular directions 100 metres above the ground level (AGL) with the mounted system. 
The trajectory calculation used Waypoint Inertial Explorer (version 8.9, NovAtel Inc., Canada) and an individual virtual GNSS base station from FINPOS service (RINEX 3.04), located approximately at the centre of the site. 
The system specifications are listed in Table~\ref{tab:scanner_details}. 

\begin{figure}
    \centering
    \includegraphics[width=\linewidth]{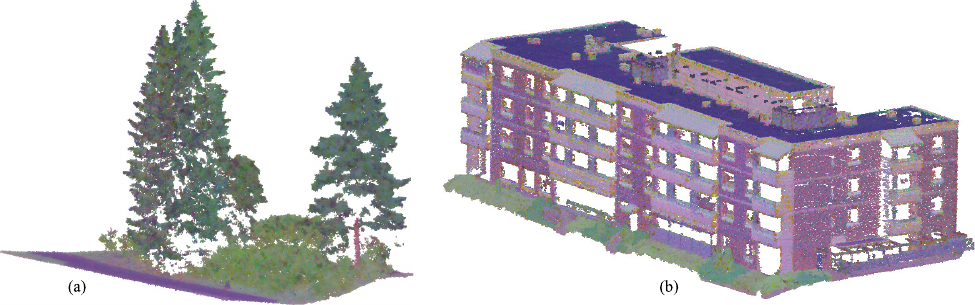}
    \caption{The high-density HeliALS data displays detailed geometric information, for example, tree branches (a) and balconies (b). The RGB colours of the figure are mapped to the reflectance values from scanners 1, 2, and 3, respectively.}
    \label{fig:closeup_pointcloud}
\end{figure}

The multispectral information consists of reflectance values provided by the Riegl scanners. The final point cloud (Section~\ref{sec:preprocessing}) has an extremely high average density of 1200 points per square metre. An illustration of the multispectral high-density data is shown in Figure~\ref{fig:closeup_pointcloud}. The scanned area marked in Figure~\ref{fig:ortophoto} holds typical urban semantic classes, mainly buildings, vegetation, and road infrastructure.

\begin{table*}[]
    \centering
    \begin{tabular}{lccc}
         Scanner&   \textbf{1}& \textbf{2}&\textbf{3}\\
         \hline
         Model& 
     VUX-1HA& miniVUX-1DL&VQ-840-G\\
 Wavelength (nm)& 1,550& 905&532\\
 Flight altitude AGL (m)& 100& 100 & 100\\
 Approx. point density (points/m$^2$)& 630& 200&420\\
 Number of returns (max.)& 9& 5& 5\\
 Maximum scanning angle ($^{\circ}$)& 360& 46*& 28 $\times$ 40\\
 Laser beam divergence (mrad)& 0.5 & 0.5 $\times$ 1.6 & 1** \\
 Laser beam diameter at the ground level (cm)& 5& 5 $\times$ 16& 10\\
 Pulse repetition rate (kHz)& 1,017 & 100 & 200\\
 Scan rate (Hz)& 143 & 72 & 100\\
 \hline
    \end{tabular}
    \caption{The specifications of the laser scanners in the multispectral laser scanning system HeliALS. The point density was calculated on the final point cloud (Section~\ref{sec:pcattr}) as the number of points per square metre. Multiple returns for example from trees can increase the point density. Scanner 2 has a circular scan pattern (*) and scanner 3 has a receiver aperture of 6 mrad (**). }
    \label{tab:scanner_details}
\end{table*}

\subsection{Dataset preprocessing}
\label{sec:preprocessing}
\begin{figure}
    \centering
    \includegraphics[width=9cm]{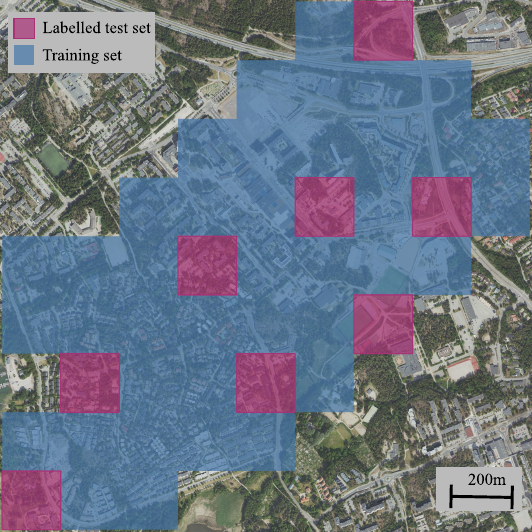}
    \caption{The scanned area is marked on the orthophoto and divided into unlabelled training and manually labelled test set primary tiles. Each tile covers a 200-metre by 200-metre area.  Photo reference: ``Orthophoto (c) Helsinki, Espoo, Vantaa, Kauniainen, Kirkkonummi, Kerava, Nurmijärvi, HSY, HSL and The Finnish Defence Forces 2023."}
    \label{fig:ortophoto}
\end{figure}

To ease the data processing, the scanned area was divided into 200-metre by 200-metre primary map tiles as shown in Figure~\ref{fig:ortophoto}. The dataset consists of 47 primary tiles, each with 42--80 million points after applying a statistical outlier removal with CloudCompare (version 2.12.4, GPL software). Eight primary tiles were selected as a separate test set (Figure~\ref{fig:ortophoto}), which was labelled and used to evaluate the methods (Section~\ref{sec_test_set}). The remaining 39 primary tiles comprised an unlabelled training set.

\begin{figure}
    \centering
    \includegraphics[width=14cm]{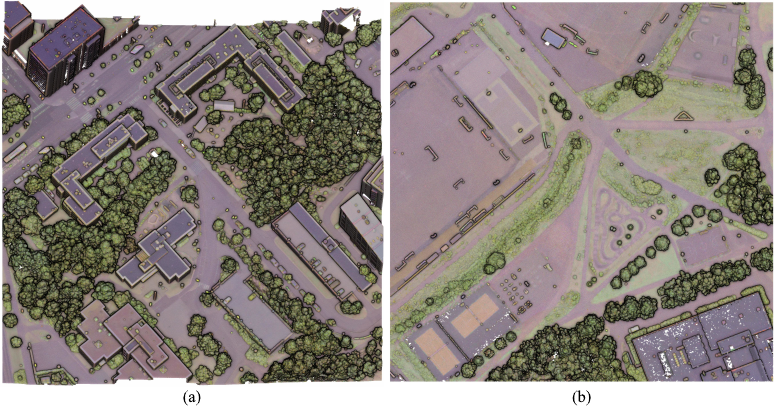}
    \caption{Two examples of 200 m \(\times\) 200 m primary tiles of urban multispectral point clouds created by combining three individual point clouds acquired with different wavelengths. Each point was assigned two additional reflectance values from the nearest neighbours acquired with the other two scanners. The RGB colours of the figure are mapped to the reflectance values from scanners 1, 2, and 3, respectively.}
    \label{fig:false_color}
\end{figure}

The multispectral point cloud was formed of three separate point clouds by searching for each point the nearest neighbour within a 25cm radius from the other two wavelengths resulting in three reflectance values for each point. Approximately 154 million points (6\%)  had missing full reflectance information and were removed as incomplete. Figure~\ref{fig:false_color} shows two examples of the multispectral point cloud with false colours.

The primary tiles were split into smaller 50-metre by 50-metre secondary tiles. For each secondary tile, the $x,y$-coordinates were centred to zero to reduce numerical errors inevitable with the original map coordinates with $x$ and $y$ of magnitude $10^5$ and $10^6$, respectively, and the $z$-coordinate was normalized by subtracting the minimum $z$ value. 

\subsubsection{Point cloud attributes}
\label{sec:pcattr}

\begin{figure}[h!]
    \centering
    \includegraphics[width =10cm]{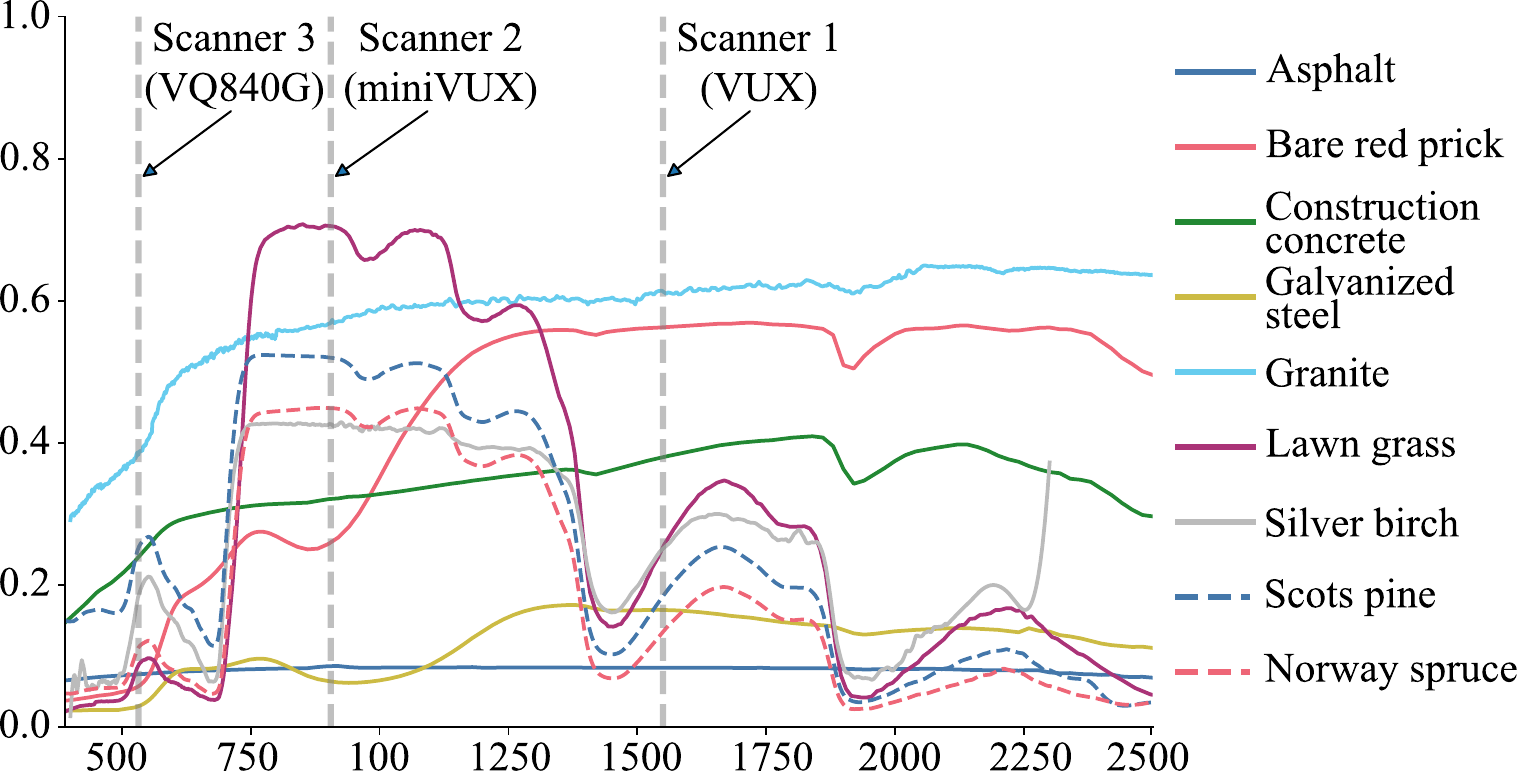}
    \caption{Characteristic reflectance spectra for various materials and objects found in the Espoonlahti multispectral HeliALS dataset. The data has been obtained from the 25 boreal tree species spectral library \cite{hovi2017spectral} and from the ASTER and the USGS spectral libraries \cite{baldridge2009aster,kokaly2017usgs}.}
    \label{fig:spectralcurves}
\end{figure}

\begin{figure}[h!]
    \centering
    \includegraphics[width=\linewidth]{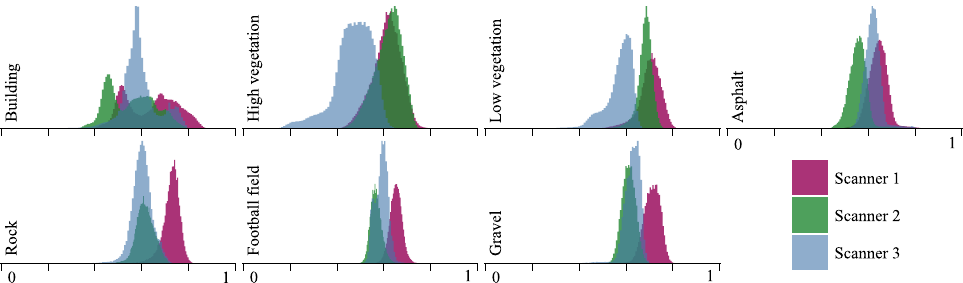}
    \caption{The distributions of the reflectance values $\rho_1, \rho_2, \rho_3$ in the test set from scanners 1,2, and 3, respectively. Reflectances were scaled between 0 and 1.}
    \label{fig:reflectance_histograms}
\end{figure}

Multispectral reflectance data contains several measurements for each point from the reflectance curve distinct for different materials as highlighted in Figure~\ref{fig:spectralcurves}. Thus, the multispectral aspect should ease distinguishing different materials. 
The distributions of reflectance values $\rho_1, \rho_2, \rho_3$ in the annotated test set classes (Section~\ref{sec_test_set}) are shown in Figure~\ref{fig:reflectance_histograms}. 

The energy of the shot laser pulse has a Gaussian form and width of one nanosecond
in the temporal dimension. However, the pulse can
stretch when reflected on non-perpendicular surfaces. Each scanner provided an echo deviation value, which measures this change between the shot and returned laser pulses. The importance of echo width in land cover classification
has been shown in several studies, for example in \cite{Qin2022AirborneClassification}.
Points with echo deviation values larger than $35$ were removed for the final point cloud. The value was set higher than the Riegl recommendation (10--15) to avoid losing too many points. As such, there were 270 million removed points, which was 10\% of the preprocessed points.
The reflectance values and echo deviation were normalized between 0 and 1.

In addition, several local geometrical and multispectral pointwise attributes were derived, including the pseudo normalized difference vegetation index PseudoNDVI$=(\rho_2-\rho_3)/(\rho_2+\rho_3)$ \citep{wichmann2015evaluating}. The linearity $(\lambda_1 - \lambda_2)/\lambda_1$, planarity $(\lambda_2-\lambda_3)/\lambda_1$, sphericity $\lambda_3 / \lambda_1$ \citep{Demantke2012DimensionalityClouds}, and verticality \citep{Guinard2017WeaklyClouds} were calculated using the eigenvalues $\lambda_1 \geq \lambda_2 \geq \lambda_3$ of the variance-covariance matrix of the nearest neighbours from 0.3m radius. 

A point's height from the ground is a useful attribute to distinguish between ground and building roofs, for example. Cloth simulation \citep{zhang2016easy} and its implementation in PDAL Python library \citep{pdal_contributors_2024} were used to extract ground points. Based on visual validation, 500 iterations were performed and a 0.5m resolution, 0.5m threshold, step size of 0.5, and rigidness of 2 were used.

\subsubsection{Superpoint forming}
\label{sec:superpoint_construction}
Each secondary tile was divided into superpoints, i.e., small homogeneous local segments (Table~\ref{tab:sp_accuracies}) for the training of the unsupervised neural network. Because the superpoints are local and lack semantic context, they cannot be considered as the semantic segmentation of the scene. On the annotated test set tiles (Figure~\ref{fig:ortophoto}), sixty-two per cent of the superpoints had a ground truth label (Section~\ref{sec_test_set}). 

An efficient superpoint construction method for large point clouds from \cite{robert2023efficient} was applied separately for ground and non-ground points. The iterative hierarchical partition process 
used an adjacency graph on spatial and spectral attributes (Table~\ref{tab:alg_point_attributes}) from 25 nearest neighbours in a one-metre range and an energy minimization problem. The attributes were chosen as in the original work. The granularity of the partition can be tuned with a regularization parameter $\gamma > 0$, where a higher value gives fewer and coarser components. 
Three iterations were run with regularization $\gamma = \{0.1,0.2,0.3\}$, weights $\{0.5,0.1,0.01\}$ to mitigate spatial information, and $\{10,50,100\}$ as the minimum for the number of points in each superpoint. The parameters were set after visual inspection and the superpoints were taken from the last level of the hierarchical partition.  When the superpoints were used in algorithms in Sections \ref{sec:growsp}--\ref{sec:random_forest}, their attributes were calculated as an average of the pointwise attributes.
 
\begin{table*}[h!]
    \centering
    \begin{tabular}{lccccccc}
                  &Building&  High veg.&  Low veg.&  Asphalt&  Rock&  Football f.&Gravel\\
         \hline         
      IoU&99.9& 99.9& 99.5& 100& 99.8& 100&97.4\\
  
 Size ($10^3$) & 34.3& 132.7& 13.9& 8.0& 0.9& 1.4&1.2\\
 Size (\%)& 17.8& 69.0& 7.2& 4.1& 0.5& 0.7&0.6\\
 \hline
    \end{tabular}
    \caption{The semantic consistency of the superpoints was examined using intersection over union (IoU) given in Section~\ref{sec:accuracy_evaluation}. The IoUs were calculated pointwise by assigning each superpoint a ground truth label by majority voting. Each secondary tile contained 2400 superpoints on average and each superpoint had 1200 points on average.}
    \label{tab:sp_accuracies}
\end{table*}

\begin{table*}[h!]
    \centering
    \setlength{\tabcolsep}{2pt}
    \begin{tabular}{lccccccc}
         & Form SPs &    GrowSP
&Cluster SPs& NN&  Predict& K-means
&RF\\
         Attribute&  Sec. \ref{sec:superpoint_construction}&  Sec. \ref{sec:growsp}&Sec. \ref{sec:growsp} ,\ref{sec:unsupervised_dl}&   Sec. \ref{sec:growsp}, \ref{sec:unsupervised_dl}& Sec. \ref{sec:unsupervised_dl}& Sec. \ref{sec:kmeans}&Sec. \ref{sec:random_forest}\\
         \hline
        Coordinates $x, y, z$&   \checkmark&    
\checkmark&-&\checkmark&    -& -
&-\\
         Reflectances $\rho_1, \rho_2, \rho_3$&   \checkmark&    \checkmark
&\checkmark&\checkmark&   \checkmark& \checkmark
&\checkmark \\
         Echo deviation&   -&    
\checkmark&\checkmark &\checkmark&    \checkmark& \checkmark
&\checkmark\\
         Linearity&   \checkmark&    \checkmark
&\checkmark &-&    \checkmark& \checkmark
&\checkmark\\
         Planarity&   \checkmark&    
\checkmark&\checkmark&-&    \checkmark& \checkmark
&\checkmark\\
 Sphericity&  \checkmark&   \checkmark
&\checkmark&-&   \checkmark& \checkmark
&\checkmark\\
 Verticality&  \checkmark&   
\checkmark&\checkmark&-&   \checkmark& \checkmark
&\checkmark\\
 Elevation& \checkmark&  -
&-& -&  -& -
&-\\
 Ground & -&  
-&-& -&  -& -
&\checkmark\\
 Neural features& -&  \checkmark&\checkmark& -&  \checkmark& -&-\\
 \hline
    \end{tabular}
    \caption{Point attributes that were used while forming the superpoints (Form SPs), growing the superpoints in the GrowSP algorithm (GrowSP), in the superpoint clustering (Cluster SPs), inputs for the neural network (NN), predicting labels for the unseen data in the GroupSP algorithm (Predict), and in reference methods (K-means and RF). The attributes were selected for superpoint forming as in \cite{robert2023efficient} and for superpoint growing / clustering and the NN following ideas from \cite{zhang2023growsp}. In the GroupSP predicting phase, the spectral and geometric attributes were noticed to improve the semantic segmentation results. The reference methods used the same attributes as the deep learning models. Initial results indicated that echo deviation was valuable for the land cover semantic segmentation. More information is given in the sections marked in each column.}
    \label{tab:alg_point_attributes}
\end{table*}

\subsection{Test set}
\label{sec_test_set}

A test set point cloud was annotated for quantitative evaluation.
Table~\ref{tab:test_set_classes} lists the test set classes and their sizes. The annotations were non-exhaustive; only points that could be easily verified to belong to the given class were selected. Thus, 52\% of the test set points were assigned a class. 

Asphalt and vegetation points were annotated from the previously classified ground points with the help of PseudoNDVI and false colours.
An unsupervised GrowSP was trained (Section~\ref{sec:growsp}) using the secondary training data tiles (Figure~\ref{fig:ortophoto}) to produce semantically consistent segments that speeded up the manual annotation process. The annotator cleaned and combined the segments that contained building and high vegetation points for each primary test set tile separately. Rock was recognized from orthophotos taken in June 2023 with 5 cm resolution and published in \cite{espooKarttapalvelu}. Gravel was verified with the help of previously collected reference points \citep{karila2023automatic} and in the field. Football fields were easily recognized based on the reflectance values.

\begin{table*}[h!]
    \centering
    \begin{tabular}{lccl}
           Class& Size ($10^6$) & Size (\%)  &Properties\\
           \hline
 Not annotated& 181& 48.2&-\\
         
           Building& 36& 9.4& Residential, public, high- and low-rise, different materials \\
           High vegetation& 97& 25.7& Various tree species (forest, gardens, along roads)\\
           Low vegetation& 26& 6.8&Grass, meadows, small bushes\\
           Asphalt& 25& 6.7&Road, parking lot\\
           Rock&   2& 0.5&Rocky areas with bare or slightly vegetated surface\\
           Football field& 6& 1.7& Areas covered with artificial turf\\
           Gravel&  4& 1.0& Soft, non-vegetated surfaces with different grain sizes\\
           &&&(small roads, sports fields, beaches)\\
           \hline
 Total & 377& 100&\\
    \end{tabular}
    \caption{The manually annotated test set consisted of 7 ground truth classes and covered 52\% of all the test set points. High vegetation was the largest and rock the smallest class.}
    \label{tab:test_set_classes}
\end{table*}

\subsection{Algorithms}
We propose a ground-aware unsupervised deep clustering method GroupSP (Section~\ref{sec:unsupervised_dl}) inspired by GrowSP (Section~\ref{sec:growsp}) to semantically segment the HeliALS dataset.
GroupSP and GrowSP were compared with K-means and supervised random forest classifier in the semantic segmentation task.
All used attributes are listed in Table~\ref{tab:alg_point_attributes}.

\subsubsection{GrowSP}
\label{sec:growsp}

\begin{figure}[h!]
    \centering
    \includegraphics[width=16cm]{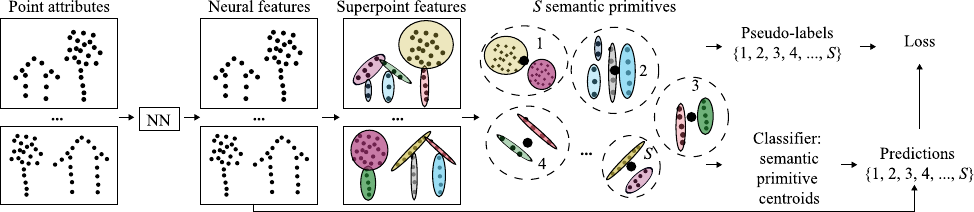}
    \caption{GrowSP computes pointwise neural features and aggregates them for precalculated superpoints. The superpoints are clustered into $S$ semantic primitives in the feature space. This gives pseudo-labels and predictions when each point is classified to the closest semantic primitive. The loss can be calculated between pseudo-labels and predictions.}
    \label{fig:growsp_chart}
\end{figure}

The fully unsupervised semantic segmentation method GrowSP proposed in \cite{zhang2023growsp} exploits deep clustering techniques: the neural network is trained iteratively by learning neural features and clustering them to produce pseudo-labels. GrowSP uses superpoints in the clustering assignment. Each superpoint likely presents a part of a semantic object and thus clustering them forms semantic primitives, i.e., an initial guess of which superpoints might belong to the same semantic segment. The idea is visualized in Figure~\ref{fig:growsp_chart}. While the training progresses, the neural features improve and the superpoints can be spatially grown based on them. 

At the preprocessing phase, GrowSP forms initial superpoints for the training data and initializes a deep neural network that takes a point cloud with $A$ attributes as input and outputs an $F$-dimensional neural feature for each point. 
In the training phase, all the superpoints from the entire training dataset are clustered in an $M+F$-dimensional feature space into $S$ semantic primitives using K-means. $M$ is the size of a given non-deep attribute set that can contain, for example, spectral and / or geometrical attributes. The semantic primitives provide pointwise pseudo-labels $\in \{1,\dots, S\}$. Pointwise predictions are made by calculating for each point the nearest semantic primitive in the neural feature space. The neural network parameters are updated based on the loss calculated between the pseudo-labels and predictions. After a predefined number of epochs, the superpoints gradually grow in spatial dimension separately for each training scene. The growing uses K-means.

In the testing phase, the neural features are calculated for each test set point. The centroids of $S$ semantic primitives are clustered into $C$ ground truth classes using K-means. Centroids of $C$ classes are used as the final classifier, i.e., a point is classified to the closest of the $C$ centroids in the neural feature space. The superpoints are no longer needed. 

The implementation was done with a desktop computer with specifications given in Table~\ref{tab:computational_hardware}.
Similarly to the original work, an efficient neural network SparseConvNet \citep{graham20183d} for naturally sparse ALS data, OneCycle learning rate \citep{smith2019super}, AdamW optimizer \citep{loshchilov2019decoupled}, and cross-entropy loss were used. The neural network initialized using Kaiming normal distribution \citep{He2015delving} took in attributes listed in Table~\ref{tab:alg_point_attributes} and produced 128-dimensional features. Echo deviation was added as an attribute because it can be useful when distinguishing land cover classes. For the neural network, data was quantized using a 15 cm voxel resolution and augmented by rotating, shifting, and scaling the coordinates.
The superpoints (Section~\ref{sec:superpoint_construction}) were clustered into new semantic primitives $S=300$ every 20th epoch. The growing was executed at epochs $e \in \{60,80,100\}$. Clustering and growing used attributes listed in Table~\ref{tab:alg_point_attributes}. For the clustering, the spectral attributes were given a weight 10. The parameters were based on initial experiments and the number of epochs $E=120$ was chosen to keep the training time reasonable (around 48 hours). Batch size 1 was chosen due to the limited computing memory. The initial tests suggested, that the semantic segmentation results improve if the number of predicted class $P$ is set higher than $C$.

\begin{table*}[h!]
    \centering
    \begin{tabular}{ll}
       Feature  & Definition  \\
       \hline
        CPU  & Intel Xeon Gold 6234 (16 physical cores @3.30GHz)\\
        RAM & 256GB (DDR4@3200MHz) \\
        GPU & Nvidia Quadro RTX 6000 (24GB of memory) \\
        Python version & 3.18.2 \\
        Cuda version & 11.3\\
        Pytorch version & 1.10.2 \\
        Operating system & Ubuntu 20.04.2 LTS \\
        \hline
    \end{tabular}
    \caption{Details of the computational hardware and software used to train the deep learning models.}
    \label{tab:computational_hardware}
\end{table*} 

\subsubsection{Ground-aware unsupervised deep clustering approach GroupSP}
\label{sec:unsupervised_dl}

\begin{figure}[h!]
    \centering
    \includegraphics[width=16cm]{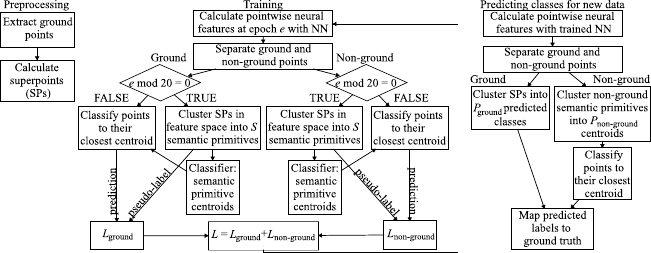}
    \caption{The ground-aware unsupervised deep clustering training pipeline GroupSP learns neural features and aggregates them using precalculated superpoints. The superpoints are clustered every 20th epoch to provide pseudo-labels and a classifier to train the network.}
    \label{fig:training_pipeline}
\end{figure}

Inspired by GrowSP, we propose a ground-aware unsupervised deep clustering method GroupSP that groups similar superpoints using multispectral information and learned deep neural features.
In comparison to GrowSP, we used the heuristic ground extraction method as a part of the training pipeline, did not iteratively grow the superpoints, and used different post-processing techniques in the testing phase, i.e., predicted $P > C$ classes which were mapped to $C$ ground truth classes.
Figure~\ref{fig:training_pipeline} shows the unsupervised training pipeline. 

Following GrowSP, GroupSP starts by calculating neural features for individual points and then divides the data into ground and non-ground points as shown in Figure~\ref{fig:training_pipeline}. The ground and non-ground superpoints are clustered (Table~\ref{tab:alg_point_attributes}) separately totalling $2S$ semantic primitives. Using the semantic primitives as pseudo-labels and their centroids as a classifier for individual points, the loss is calculated as $L = L_{\text{ground}}+L_{\text{non-ground}}$. Based on initial tests, the model was trained for $E=80$ epochs, and the semantic primitive clustering was done at epochs $e \in \{ 20, 40, 60\}$ such that the spectral attributes of the ground superpoints were given weight 10. The number of semantic primitives was set to $S=300$. For further training details see Section~\ref{sec:growsp}.

\textbf{Predicting classes} for the new data follows the idea from \cite{de2023dc3dcd} where the test set data is oversegmented and the predicted classes are mapped to ground truth classes. The mapping allows the user to define interesting ground truth classes but does not require that the scene should contain only instances from them. This is especially important with urban scenes, which involve unseen objects, several semantic classes and subclasses (e.g., tree species) and great variation inside semantic classes (e.g., small buildings and high commercial buildings). We calculated pointwise neural features for the test set and separated ground and non-ground points. For the non-ground points, a classifier was formed as in GrowSP by clustering the centroids of the $S=300$ non-ground semantic primitives into $P_{\text{non-ground}}=50$ centroids. A point belongs to the closest one in the neural feature space. For the ground points, initial experiments suggested that a similar classifier did not provide good results on the smallest ground truth classes. Thus, we chose a computationally more demanding approach which was also used in the pseudo-label forming: the superpoints (Section~\ref{sec:superpoint_construction}) were clustered with K-means in the feature space (Table~\ref{tab:alg_point_attributes}) into $P_{\text{ground}}=300$ predicted classes. Superpoints were chosen instead of points for faster clustering. 
$P_{\text{ground}}$ and $P_{\text{non-ground}}$ were set based on initial experiments.

\subsubsection{Unsupervised reference semantic segmentation method}
\label{sec:kmeans}

K-means \citep{lloyd1982least} was used as an unsupervised reference clustering method to semantically segment the HeliALS dataset. The clustering was done as in the GroupSP training phase without the neural features, to compare if the deep feature generator added any value to the semantic segmentation. For a fair comparison, the clustering was done separately for ground and non-ground superpoints (Section~\ref{sec:superpoint_construction}) with the same attributes (Table~\ref{tab:alg_point_attributes}) and weights as in Section~\ref{sec:unsupervised_dl}. Ground and non-gound points were oversegmented into 300 and 50 predicted classes, respectively. 
K-means does not require training data. Thus, only the primary test set tiles marked in Figure~\ref{fig:ortophoto} were used. 

\subsubsection{Supervised reference semantic segmentation method}
\label{sec:random_forest}

The random forest classifier \citep{breiman2001random} was chosen to provide fully supervised reference results for the unsupervised semantic segmentation methods.
RF has been applied in multispectral land cover classification, for example, in \cite{matikainen2017object}.
Instead of points, we used the RF to classify superpoints (Section~\ref{sec:superpoint_construction}). The attributes were chosen to provide RF with the same information as in GroupSP (Table~\ref{tab:alg_point_attributes}). RF used only annotated samples and stratified five-fold cross-validation. 
The class imbalance was addressed by undersampling the building and high vegetation classes to have $N/7$ samples and oversampling the other to $N/7$ samples where $N = 153963$ is the size of the training set at each fold. 
The implementation used the Python library Skicit-learn \citep{scikit-learn} version 1.2.2 and the default parameters, i.e., 100 trees whose nodes were expanded until all leaves were pure or contained less than two samples.

\subsection{Point attribute ablation study}
\label{sec:ablation}

To examine the role of the multispectral information in the semantic segmentation task of the HeliALS dataset, we conducted a point attribute ablation study. The experiments are given in Table~\ref{tab:experiments}. Similarly to \cite{hakula2023individual}, we incrementally added more spectral channels (ch) in the training pipeline; the superpoint forming, neural network, and predicting classes for new data used only the available spectral attributes. In all of the experiments, the geometries from all scanners were used to maintain consistency in the results. Scanner 1 was chosen as the primary as it provided the most accurate geometry and highest point density. The reflectance value from scanner 2 was added as a second because it was acquired 
on near-infrared wavelength typical in ALS.

\begin{table}[h!]
    \centering
    \small
    \begin{tabular}{lccccc}
     & & & & & \\
         Experiment&       Coordinates&$\rho_1$&$\rho_2$&$\rho_3$&Echo deviation\\
         \hline
 Geometry&     \checkmark&-&-&-&-\\
 One spectral ch &     \checkmark&\checkmark&-&-&-\\
 Two spectral chs&     \checkmark&\checkmark&\checkmark&-&-\\
 Three spectral chs&  \checkmark&\checkmark& \checkmark& \checkmark& -\\
 Full model&  \checkmark&\checkmark& \checkmark& \checkmark& \checkmark\\
 \hline
 \end{tabular}
    \caption{Point attributes fed in the neural network in the multispectral information ablation study. The reflectances \(\rho_1\), \(\rho_2\), and \(\rho_3\) as well as the echo deviation are described in Section~\ref{sec:pcattr}.}
    \label{tab:experiments}
\end{table}

\subsection{Accuracy evaluation}
\label{sec:accuracy_evaluation}
The accuracies of the models were evaluated using the annotated test set point cloud (Section~\ref{sec_test_set}) and typical semantic segmentation metrics \citep{zhang2023deep-learning}, i.e., overall accuracy $\text{oAcc}  = 1/n \sum ^C _{i=1}w_{ii}$, mean accuracy $\text{mAcc}  = 1/C \sum ^C _{i=1} (w_{ii}/\sum ^C _{j=1} w_{ij})$, and mean intersection over union $\text{mIoU}  = 1/C \sum ^C _{i=1} \text{IoU}_i$, where ${\text{IoU}_i =  w_{ii}/(\sum ^C _{j=1} w_{ij} + \sum^C_{k=1}w_{ki}-w_{ii})}$, $w_{ij}$ is the number of points that belong to class $i$ and are classified to class $j$, $n$ is the number of points and $C$ number of classes.

In unsupervised semantic segmentation, the predicted classes $P\geq C$ must be mapped to ground truth classes. We used majority voting which assigned each predicted class a ground truth label based on their pointwise annotations. The mapped classes did not necessarily hold all $C$ ground truth classes and predicted classes without annotations were left unmapped. All the accuracy metrics were calculated pointwise.
The evaluation of the supervised RF classifier (Section~\ref{sec:random_forest}) was done by calculating the average of the evaluation metrics over the five folds. 

\section{Results and discussion}
\label{sec:resultsanddiscussion}

The proposed GroupSP algorithm is examined in the semantic segmentation task of the multispectral HeliALS data in Section~\ref{sec:novel_results}. GroupSP is compared to three other models, i.e., GrowSP, K-means, and supervised random forest in Section~\ref{sec:compare_models}, and Section~\ref{sec:ablationResults} presents an ablation study to showcase the importance of multispectral information.

\subsection{Assessing GroupSP}
\label{sec:novel_results}

A confusion matrix for GroupSP is given in Table~\ref{tab:conf_matrix}.
Additionally, GroupSP was evaluated in three experiments (Table~\ref{tab:groupsp_exp}). First, the predicted classes were mapped to ground truth classes using $N$ randomly chosen annotated samples per ground truth class. A mIoU of 75\% was reached with  $N=1000$, i.e., 7000 annotated points, which is 0.004\% of all the available annotations. In theory, producing $N=1000$ is straightforward, for example, in the high-density HeliALS dataset a single tree can contain over ten thousand points. However, in practice, it would be reasonable to choose the samples from diverse instances across the test set to ensure comprehensive coverage. High vegetation suffered the least and rock the most from the reduced annotations. 

Second, the number of predicted classes $P_{\text{ground}}$ and $P_{\text{non-ground}}$ were varied. $P_{\text{ground}}$ needed to be sufficiently high not to merge the smaller land cover classes with the larger ones. The non-ground classes were more robust toward  $P_{\text{non-ground}}$ possibly because both were easily distinguishable in geometry.

Lastly, the accuracies were compared across the number of training epochs $E$ up to 140. The results improved clearly at $E=40$, largely because of increased rock, football field, and gravel accuracies. When $E> 80$, only the gravel class accuracy improved. The accuracy and number of training epochs did not correlate because nothing guided the unsupervised network toward the given ground truth classes when the training progressed; the most distinctive features for our test set were obtained at $E=80$. In the future, the neural network could be assisted, for example, with active \citep{lin2020active} or few-shot learning \citep{huang2024simple}.

Furthermore, only annotating the points that were easy to recognise, i.e., 52\% of the entire test set area, could have introduced inherent bias in the accuracy evaluation: the easiest points to label are likely the easiest to segment. As such, the reported accuracy might overestimate the true accuracy given data with 100\% of the points annotated.

\begin{table*}[h!]
    \centering
    \setlength{\tabcolsep}{2pt}
    \begin{tabular}{llccccccccc}
 & & \multicolumn{3}{l}{Predicted classes ($10^6$)} & & & &  &Total ($10^6$)&Recall (\%)\\
 \cline{3-9}
          &&  Building&  High veg.&  Low veg.&  Asphalt&  Rock&  Football field& Gravel  &&\\
          \hline
          Ground &Building&  34.9&  0.3&  0.1&  0&  0&  0& 0 &35.3&98.5\\
          truth&High veg.&  0.1&  96.3&  0.1&  0&  0&  0& 0 &96.5&99.8\\
         ($10^6$) &Low veg.&  0.1&  0.9&  24.2&  0.1&  0.1&  0& 0.3 &25.7&94.3\\
          &Asphalt&  0.1&  0&  0.2&  24.3&  0&  0.3& 0.5 &25.4&96.1\\
          &Rock&  0.2&  0&  0.3&  0&  1.2&  0& 0.1 &1.8&66.7\\
          &Football field&  0&  0&  0&  0.7&  0.1&  5.4& 0.1 &6.3&86.7\\
          &Gravel&  0&  0&  0.5&  0.3&  0.2&  0.3& 2.4 &3.7&65.0\\
          \cline{2-9}
 &Total ($10^6$)& 35.4& 97.5& 25.4& 25.4& 1.6& 6.0& 3.4& &\\
 \cline{2-9}
 \multicolumn{2}{l}{Precision (\%)}& 98.8& 98.8& 94.9& 95.9& 73.5& 90.6& 70.8 &&\\
 \hline
    \end{tabular}
    \caption{A pointwise confusion matrix with classwise precision and recall for GroupSP (full model) in the semantic segmentation task of the HeliALS dataset.}
    \label{tab:conf_matrix}
\end{table*}

\begin{table*}
    \centering
    \setlength{\tabcolsep}{2pt}
    \begin{tabular}{ccccccccccc}
 & & & & & & & IoU(\%)& & &\\
 \cline{5-11}
Experiment & oAcc(\%)& mAcc(\%)& mIoU(\%)& Building & High veg. & Low veg. & Asphalt & Rock & Football f. &Gravel \\
\hline
\hline
  & & & & & & & & & &\\
 $N$ $(10^3)$ & & & & & & & & & &\\
\hline
$0.01$&  58.4&  49.7&  35.2&  30.1&  63.1&  12.2&  39.0&  18.4& 49.4&33.9\\
$0.1$&  89.8&  82.0&  66.9&  72.2&  95.3&  55.7&  89.4&  38.6& 74.2&43.0\\
$1$&  94.8&  89.2&  74.5&  92.4&  98.1&  81.1&  91.0&  38.8& 78.1&42.3\\
$10$&  95.4&  89.9&  75.3&  95.6&  98.5&  83.7&  90.5&  39.2& 74.8&44.9\\
$100$&  95.4&  89.9&  75.2&  95.5&  98.5&  83.7&  90.5&  39.2& 74.4&44.9\\
$1000$&  95.4&  89.9&  75.3&  95.6&  98.5&  83.7&  90.4&  39.2& 74.8&44.8\\
  & & & & & & & & & &\\
$P_{\text{ground}}, P_{\text{non-ground}}$    & & & & & & & & & &\\
\hline
200, 50&  96.8&  85.4&  79.4&  97.0&  98.5&  89.5&  91.7&  50.6&  80.0&48.6\\
100, 50&  96.4&  81.4&  75.6&  96.8&  98.5&  88.4&  90.4&  38.9&  76.3&39.7\\
50, 50&  95.4&  77.8&  72.0&  96.3&  98.5&  86.4&  83.9&  36.7&  62.2&40.1\\
25, 50&  93.1&  73.4&  65.5&  96.3&  98.5&  76.6&  72.0&  33.2&  49.3&32.7\\
300, 100& 97.0& 86.7& 80.4& 97.2& 98.6& 90.0& 92.3& 53.8& 79.5&51.3\\
300, 25&  96.8&  86.6&  80.2&  96.8&  98.3&  89.7&  92.3&  53.8&  79.5&51.3\\
300, 10&  96.4&  86.6&  79.8&  94.6&  97.5&  89.7&  92.3&  53.8&  79.5&51.3\\
  & & & & & & & & & &\\
 $E$   & & & & & & & & & &\\
\hline
         20&  94.9&  74.4&  67.9&  95.1&  97.5&  83.0&  89.3&  8.0&  72.1&30.1\\
         40&  96.5&  82.2&  76.7&  96.9&  98.4&  86.6&  92.9&  34.1&  85.7&42.5\\
         60&  96.6&  83.8&  77.3&  97.2&  98.5&  88.4&  91.7&  43.5&  83.5&38.4\\
         
         100&  96.8&  84.6&  78.9&  97.0&  98.5&  89.6&  91.7&  44.4&  77.3&54.0\\
 120& 96.8& 84.5& 78.9& 96.4& 98.6& 88.1& 92.0& 39.6& 83.5&54.4\\
 140& 97.1& 86.6& 80.9& 96.9& 98.7& 90.2& 93.0& 47.6& 83.6&56.0\\
  & & & & & & & & & &\\
Full model   & & & & & & & & & &\\
\hline
all / 300, 50 / 80 &  96.9&  86.7&  80.3&  97.3&  98.5&  89.7&  92.3&  53.8& 79.5&51.3\\
\hline 
\hline
    \end{tabular}
    \caption{The performance of the GroupSP algorithm when the number of annotated points $N$ per ground truth class used in the mapping assignment, the number of predicted classes $P_{\text{ground}}, P_{\text{non-ground}}$, and the number of training epochs $E$ were varied. The full model used all the annotated points in the mapping assignment (Table~\ref{tab:test_set_classes}), divided the ground points into $P_{\text{ground}} = 300$ and the non-ground points into $P_{\text{non-ground}} = 50$ classes, and trained the neural network 80 epochs.}
    \label{tab:groupsp_exp}
\end{table*}

\subsection{Comparison to other semantic segmentation methods}
\label{sec:compare_models}

\begin{figure*}[h!]
    \centering
    \includegraphics[width=\textwidth]{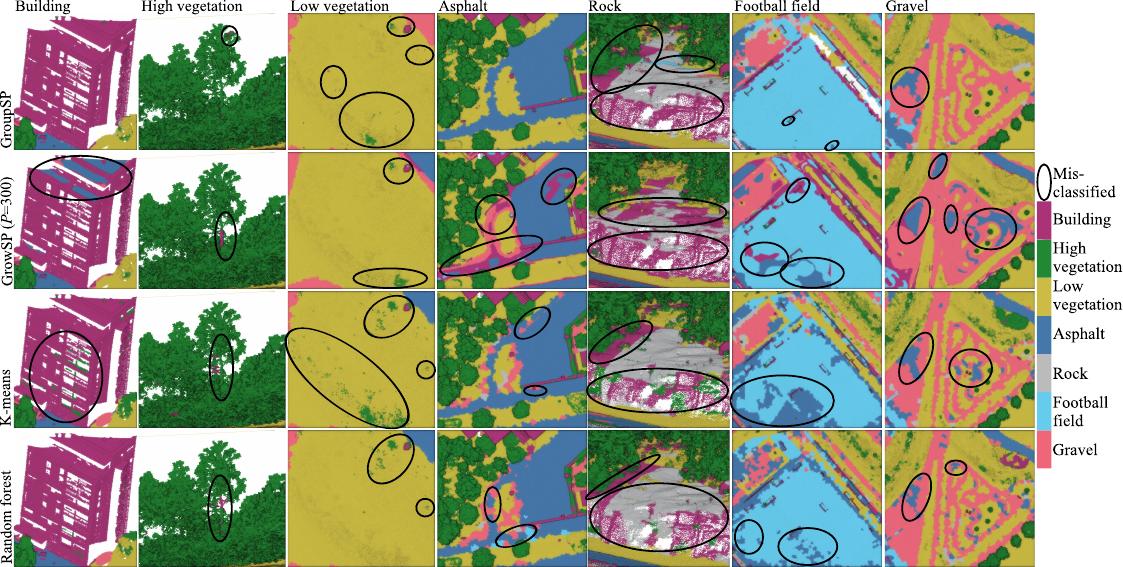}
    \caption{Comparison of the unsupervised GroupSP, GrowSP, K-means, and supervised RF performance on the ground truth classes.}
    \label{fig:gt_results}
\end{figure*}

\begin{figure*}
    \centering
    \includegraphics[width=14cm]{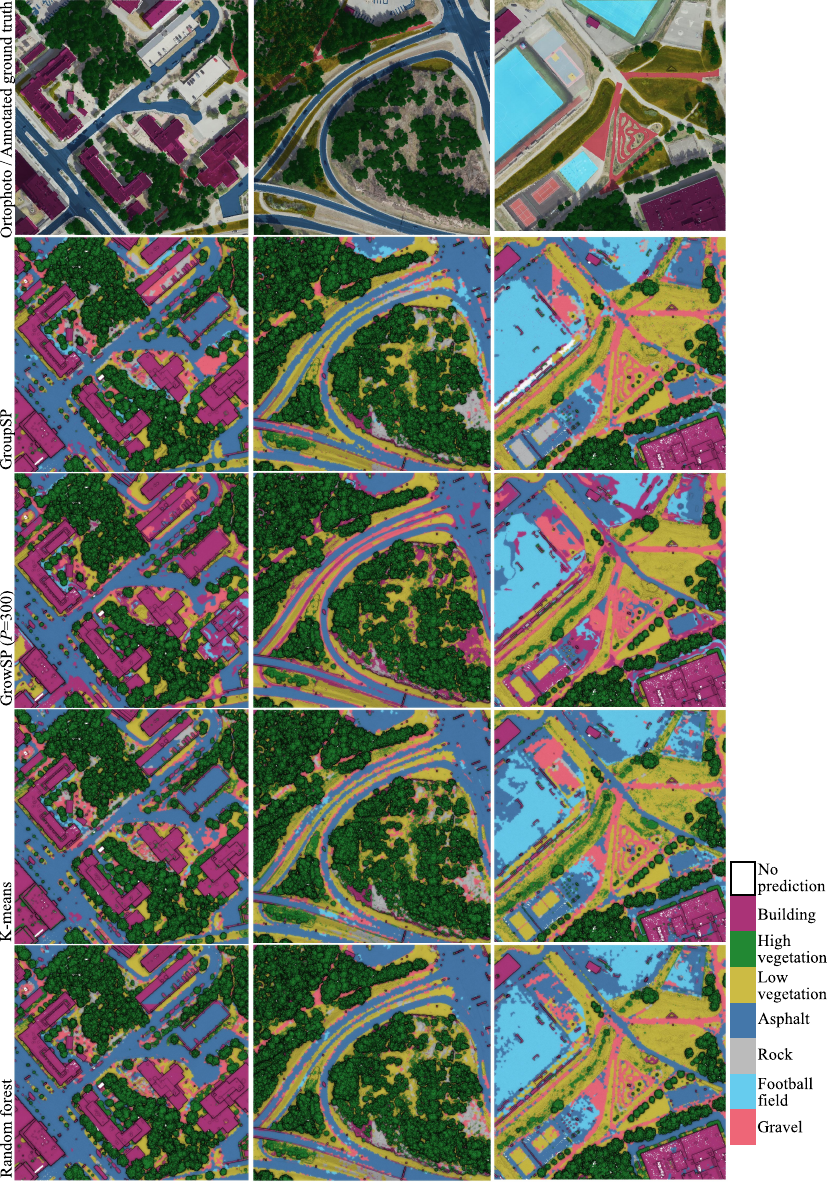}
    \caption{Visualization of the semantic segmentation of HeliALS dataset with unsupervised GroupSP, GrowSP, K-means, and supervised RF. Photo reference: ``Orthophoto (c) Helsinki, Espoo, Vantaa, Kauniainen, Kirkkonummi, Kerava, Nurmijärvi, HSY, HSL and The Finnish Defence Forces 2023."}
    \label{fig:overall_scene_results}
\end{figure*}

Table~\ref{tab:model_comparison_results} presents a quantitative comparison of the semantic segmentation results of unsupervised GroupSP, GrowSP, K-means, and supervised random forest classifier. Figure~\ref{fig:gt_results} compares the performances in the ground truth classes qualitatively and Figure~\ref{fig:overall_scene_results} shows the results on entire primary test set tiles. The most significant differences between the models were seen with the smallest classes. 

\begin{table*}[h!]
    \centering
    \setlength{\tabcolsep}{2pt}
    \begin{tabular}{lcccccccccc}
 & & & & & & & IoU(\%)& & &\\
  \cline{5-11}
         Experiment&  oAcc(\%)&  mAcc(\%)&  mIoU(\%)&  Building & High veg. &  Low veg. &  Asphalt &  Rock &  Football f. & Gravel \\
          \hline
 GroupSP & 96.9& 86.7& 80.3& 97.3& 98.5& 89.7& 92.3& 53.8& 79.5&51.3\\
 GrowSP ($P$=7)& 59.9& 36.6& 25.5& 19.0& 73.2& 36.7& 46.5& 3.0& 0&0\\
 GrowSP ($P$=300)& 92.2& 71.7& 63.0& 76.2& 97.4& 81.9& 84.6& 9.9& 56.4&34.5\\
        
         K-means & 95.3& 83.9&75.7& 93.3& 97.4& 91.4& 83.8& 51.7& 59.8& 52.2\\ 
         Random forest & 97.7& 91.5& 86.2& 96.9& 99.0& 94.2& 91.8& 71.1& 82.5& 67.7\\
         
         \hline
    \end{tabular}
    \caption{Quantitative accuracy metrics for unsupervised GroupSP, GrowSP, K-means, and supervised RF in the semantic segmentation task of the HeliALS dataset.}
    \label{tab:model_comparison_results}
\end{table*}

The highest accuracy was obtained with the supervised RF, which vastly outperformed the other models in rock and gravel segmentation.
The second-best model was the unsupervised GroupSP. GroupSP was more accurate on buildings and asphalt than RF and performed almost equally on high vegetation and football field segmentation. RF might have overestimated the semantic segmentation results as the training and test sets were not separated spatially \citep{beigaite2022spatial}. However, the performance could have been underestimated if entire annotated tiles were left for the test set because none of them contained annotations from all ground truth classes, for example, only one contained football fields. 

GrowSP failed when the number of predicted classes was low $P=7$. This was expected with the highly imbalanced test set and unsupervised model trained without information on the ground truth classes. However, the segmentation accuracy improved when the number of predicted classes was increased to $P=300$. With $P=300$, GrowSP failed to separate ground from building roofs (Figure~\ref{fig:gt_results}) and GroupSP outperformed GrowSP in every evaluation metric. Thus, the heuristic ground extraction eased the segmentation.

The non-deep K-means performed almost equally to the GroupSP with high vegetation and rock and excelled GroupSP with low vegetation and gravel.  The results would suggest that learning deep features can enhance the segmentation accuracies but that good results are also possible without long training times and large training sets.

The only non-ground classes, i.e., buildings and high vegetation had distinctive properties. Thus, the building and high vegetation points were segmented accurately by most of the models. Learning neural features (GroupSP) improved the segmentation accuracy of buildings in comparison to K-means and RF. K-means misclassified many building points as high vegetation and  GroupSP the solitary non-ground points as buildings (Figure~\ref{fig:gt_results}).

The ground extraction algorithm was inconsistent with low vegetation points (e.g., shrubs). Thus, the final high and low vegetation segments had no clear boundaries. 
Because asphalt and football fields had similar spectral properties (Figure~\ref{fig:reflectance_histograms}), all models misclassified asphalt points as football fields and vice versa.
All models struggled to segment rock and gravel. 
The rock was often misclassified as a building, especially on vertical cliffs where the ground extraction algorithm had failed (Figure~\ref{fig:gt_results}).

Our results are aligned with the literature.
\cite{matikainen2017object} conducted land cover classification in the same area. Similarly to our results, they found that gravel and rock were the most challenging classes among buildings, trees, asphalt, gravel, rocky areas, and low vegetation.
The ISPRS ALS benchmark dataset \citep{niemeyer2014contextual} includes several similar classes to our test set: powerline, low vegetation, impervious surface, car, fence, roof, facade, shrub, and tree. For example, \cite{lin2022weakly}, \cite{huang2024simple}, and \cite{zhang2024havana} reported the highest accuracies in impervious surfaces, roofs, and trees while using reduced annotation effort techniques. \cite{zhang2024havana} observed, that the major classes were easier to classify correctly, while classes with similar geometric properties (e.g., fences and low vegetation) had limited improvements in accuracy. Additionally, \cite{zhang2021unsupervised} has reported similar findings of improved accuracy with unsupervised deep clustering methods.

\subsection{Point attribute ablation study results}
\label{sec:ablationResults}

\begin{figure*}[h!]
    \centering
    \includegraphics[width=\textwidth]{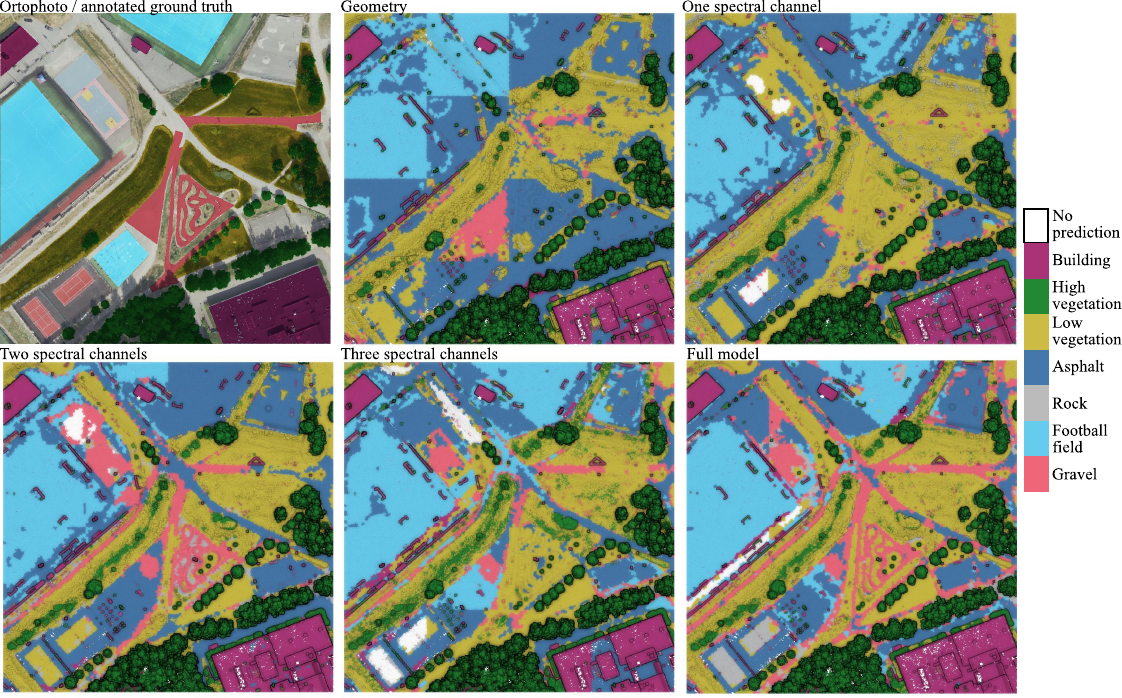}
    \caption{Visualization of the semantic segmentation of HeliALS dataset with GroupSP in the point attribute ablation study. Photo reference: ``Orthophoto (c) Helsinki, Espoo, Vantaa, Kauniainen, Kirkkonummi, Kerava, Nurmijärvi, HSY, HSL and The Finnish Defence Forces 2023."}
    \label{fig:ablation_visu}
\end{figure*}

The experiments from Table~\ref{tab:experiments} were conducted to demonstrate the importance of multispectral information in semantic segmentation. The quantitative accuracy metrics are listed in Table~\ref{tab:ablation_results} and the results are visualized in Figure~\ref{fig:ablation_visu}. Increasing the number of spectral channels from none to three improved the mIoU by 17 percentage points and from one to three by 9. The classwise accuracies did not increase consistently with the added spectral information. However, the experiments between pure geometry and three spectral channels indicate that the multispectral information is more reliable than additions from single scanners, especially on the ground classes. The echo deviation was important in distinguishing the ground classes and increased the mIoU by 8 percentage points. 

The non-ground classes, building and high vegetation, were segmented accurately without spectral information as they had distinctive geometries. Similarly, low vegetation and asphalt had higher segmentation accuracies without spectral information than the other ground classes. However, adding spectral information improved their accuracy vastly.
Rock and gravel were hard to segment, and their accuracies did not improve consistently with the added spectral channels. Echo deviation was an important attribute for both. 
The accuracy of the football fields improved most after the third spectral channel.

\begin{table*}[h!]
    \centering
    \setlength{\tabcolsep}{2pt}
    \begin{tabular}{lcccccccccc}
 & & & & & & & IoU(\%)& & &\\
  \cline{5-11}
         Experiment&  oAcc(\%)&  mAcc(\%)&  mIoU(\%)&  Building & High veg. &  Low veg. &  Asphalt &  Rock &  Football f. & Gravel \\
          \hline
 Geometry& 89.4& 63.1& 55.4& 94.6& 98.3& 56.5& 58.9& 15.6& 49.8&13.8\\
 One spectral ch& 94.0& 68.9& 63.7& 95.0& 97.9& 76.3& 84.7& 17.0& 67.6&7.2\\
 Two spectral chs& 95.2& 77.3& 71.1& 95.9& 98.2& 85.7& 84.2& 36.9& 61.8&34.9\\
 Three spectral chs& 96.0& 77.3&72.7& 96.5& 98.6& 82.6& 91.6& 30.1& 84.6& 24.8\\ 
 Full model& 96.9& 86.7& 80.3& 97.3& 98.5& 89.7& 92.3& 53.8& 79.5& 51.3\\
         
         \hline
    \end{tabular}
    \caption{Quantitative accuracy metrics of the ablation study (Sec. \ref{sec:ablation}, Table\ref{tab:experiments}) in the semantic segmentation task of the HeliALS dataset with GroupSP.}
    \label{tab:ablation_results}
\end{table*}

\section{Conclusions}
\label{sec:conclusions}
This paper introduces a novel unsupervised deep learning approach for the semantic segmentation of high-density multispectral ALS data collected from urban environments. As the semantic segmentation method, we proposed GroupSP, inspired by the GrowSP algorithm. GroupSP is a ground-aware deep clustering method that divides the scene into superpoints and groups them using neural features. The clustering assignment is used as pseudo-labels when training a neural network. Predicting labels for unseen data was done by over-segmenting the test set and mapping the predicted classes to ground truth classes using an automated mapping assignment. We assessed and compared GroupSP to GrowSP, K-means, and supervised random forest classifier with seven ground truth classes: building, high vegetation, low vegetation, asphalt, rock, football field, and gravel. A point attribute ablation study was conducted on the multispectral attributes. 

Even with a limited number of annotated points in the mapping assignment (0.004\% of the annotations), we achieved a mIoU of 75\% with the unsupervised GroupSP. However, the best performance, with a mIoU of 80\%, was obtained using all of the annotated points.

Of the compared semantic segmentation methods, the supervised random forest performed best (mIou=86\%). GroupSP was the second-best model and more accurate than GrowSP (mIoU=63\%) in every ground truth class. The heuristic ground extraction helped to prevent confusion between the building roofs and ground points. K-means (mIoU=76\%) performed almost equally to GroupSP. However, learning neural features improved the segmentation results.

The point attribute ablation study showed that adding spectral information to the model can improve the semantic segmentation results. Adding three spectral channels in comparison to none increased the mIoU by 17 percentage points and adding three compared to one by 9. Additionally, the echo deviation was important on the ground classes and increased the mIoU by 8 percentage points.

Future research could focus on enhancing the accuracy of semantic segmentation for smaller ground truth classes. This might involve incorporating a few annotated samples during training or fine-tuning the model weights. Additionally, developing methods to identify points that do not belong to any ground truth class would enable automatic filtering of point clouds, focusing solely on relevant semantic information. Further exploration of deep learning techniques for processing large point clouds is essential. For example, the training time could be reduced using superpoints as neural network inputs as in \cite{robert2023efficient}.
 
\section*{Author contributions}
\label{sec:contributions}
\textbf{Oona Oinonen} acted as the first author, wrote most of the original draft and had the main responsibility for method conceptualization and development. OO processed the data, contributed to the software, computed, analysed, validated, and visualized the results.
\textbf{Lassi Ruoppa} provided the original software and participated in developing the methodology. \textbf{Josef Taher}, \textbf{Matti Lehtomäki}, and \textbf{Leena Matikainen} were the supervisors of the work and participated in the conceptualization and development of the methodology. JT confirmed test set points resources in the field, visualized Figure~\ref{fig:spectralcurves}, and participated in writing Section~\ref{sec:multispectral_laser_scannning}. LM participated in writing the introduction and Section~\ref{sec:multispectral_laser_scannning}. \textbf{Kirsi Karila} and LM provided reference point data from the study area. \textbf{Teemu Hakala} was the flight operator while collecting the HeliALS dataset, participated in data curation, and provided Figure~\ref{fig:lidar_setup}. \textbf{Antero Kukko} acted as the project administrator and participated in the conceptualization, built the HeliALS laser scanning system and calculated the point clouds. \textbf{Harri Kaartinen} did the flight plan and helped install the HeliALS system. AK and HK participated in writing Section~\ref{sec:dataset_collection}. AK, HK, and \textbf{Juha Hyyppä} acquired the funding.
LR, JT, ML, LM, KK, TH, AK, HK, and JH reviewed \& edited the text.

\section*{Acknowledgements}
\label{sec:acknowledgements}
We gratefully acknowledge the Research Council of Finland projects ``Forest-Human-Machine Interplay -- Building Resilience, Redefining Value Networks and Enabling Meaningful Experiences" (decision no. 357908), ``Mapping of forest health, species and forest fire risks using Novel ICT Data and Approaches" (decision no. 344755), ``Understanding Wood Density Variation Within and Between Trees Using Multispectral Point Cloud Technologies and X-ray microdensitometry" (decision no. 331708), ``Capturing structural and functional diversity of trees and tree communities for supporting sustainable use of forests" (decision no. 348644), and the Ministry of Agriculture and Forestry grant ``Future Forest Information System at Individual Tree Level'' (VN/3482/2021).

\newpage
\bibliographystyle{abbrvnat}

\end{document}